\pdfoutput=1
\documentclass[10pt]{article}
\usepackage[utf8]{inputenc}
\usepackage[T1]{fontenc}

\usepackage[verbose=true,letterpaper]{geometry}
\AtBeginDocument{\newgeometry{textheight=9in,textwidth=5.5in,top=1in,headheight=12pt,headsep=25pt,footskip=30pt}}
\usepackage{amsfonts,amsmath,amssymb}
\usepackage{graphicx}
\graphicspath{{submission/}{./}}
\usepackage{float}
\usepackage{microtype}
\usepackage{url}
\usepackage{booktabs,longtable,array}
\usepackage{hyperref}
\newcommand{\citelink}[2]{\begingroup\hypersetup{linkbordercolor={0 1 0}}\hyperlink{#1}{#2}\endgroup}
\hypersetup{pdftitle={Toward a Compositional Knowledge Architecture for a Shared Organizational Cortex},pdfauthor={Andrew Miracle, Phyl Georgiou, Christine Okoth, Ese Monday}}
\input{glyphtounicode}
\makeatletter
\renewcommand{\normalsize}{%
  \@setfontsize\normalsize\@xpt\@xipt
  \abovedisplayskip      7\p@ \@plus 2\p@ \@minus 5\p@
  \abovedisplayshortskip \z@ \@plus 3\p@
  \belowdisplayskip      \abovedisplayskip
  \belowdisplayshortskip 4\p@ \@plus 3\p@ \@minus 3\p@
}
\normalsize
\renewcommand{\small}{%
  \@setfontsize\small\@ixpt\@xpt
  \abovedisplayskip      6\p@ \@plus 1.5\p@ \@minus 4\p@
  \abovedisplayshortskip \z@  \@plus 2\p@
  \belowdisplayskip      \abovedisplayskip
  \belowdisplayshortskip 3\p@ \@plus 2\p@   \@minus 2\p@
}
\renewcommand{\footnotesize}{\@setfontsize\footnotesize\@ixpt\@xpt}
\renewcommand{\scriptsize}{\@setfontsize\scriptsize\@viipt\@viiipt}
\renewcommand{\tiny}{\@setfontsize\tiny\@vipt\@viipt}
\renewcommand{\large}{\@setfontsize\large\@xiipt{14}}
\renewcommand{\Large}{\@setfontsize\Large\@xivpt{16}}
\renewcommand{\LARGE}{\@setfontsize\LARGE\@xviipt{20}}
\renewcommand{\huge}{\@setfontsize\huge\@xxpt{23}}
\renewcommand{\Huge}{\@setfontsize\Huge\@xxvpt{28}}

\renewcommand{\section}{%
  \@startsection{section}{1}{\z@}%
                {-2.0ex \@plus -0.5ex \@minus -0.2ex}%
                { 1.5ex \@plus  0.3ex \@minus  0.2ex}%
                {\large\bf\raggedright}%
}
\renewcommand{\subsection}{%
  \@startsection{subsection}{2}{\z@}%
                {-1.8ex \@plus -0.5ex \@minus -0.2ex}%
                { 0.8ex \@plus  0.2ex}%
                {\normalsize\bf\raggedright}%
}
\renewcommand{\subsubsection}{%
  \@startsection{subsubsection}{3}{\z@}%
                {-1.5ex \@plus -0.5ex \@minus -0.2ex}%
                { 0.5ex \@plus  0.2ex}%
                {\normalsize\bf\raggedright}%
}
\renewcommand{\paragraph}{%
  \@startsection{paragraph}{4}{\z@}%
                {1.5ex \@plus 0.5ex \@minus 0.2ex}%
                {-1em}%
                {\normalsize\bf}%
}

\renewcommand{\footnoterule}{\kern-3\p@ \hrule width 12pc \kern 2.6\p@}
\def\@listi  {\leftmargin\leftmargini}
\def\@listii {\leftmargin\leftmarginii
              \labelwidth\leftmarginii
              \advance\labelwidth-\labelsep
              \topsep  2\p@ \@plus 1\p@    \@minus 0.5\p@
              \parsep  1\p@ \@plus 0.5\p@ \@minus 0.5\p@
              \itemsep \parsep}
\def\@listiii{\leftmargin\leftmarginiii
              \labelwidth\leftmarginiii
              \advance\labelwidth-\labelsep
              \topsep    1\p@ \@plus 0.5\p@ \@minus 0.5\p@
              \parsep    \z@
              \partopsep 0.5\p@ \@plus 0\p@ \@minus 0.5\p@
              \itemsep \topsep}

\newcommand{\@toptitlebar}{%
  \hrule height 4\p@
  \vskip 0.25in
  \vskip -\parskip%
}
\newcommand{\@bottomtitlebar}{%
  \vskip 0.29in
  \vskip -\parskip
  \hrule height 1\p@
  \vskip 0.09in%
}
\newcommand{\ftype@noticebox}{8}
\newcommand{\@notice}{%
  \enlargethispage{2\baselineskip}%
  \@float{noticebox}[b]%
    \footnotesize Preprint. November 2026.%
  \end@float%
}
\renewcommand{\maketitle}{%
  \par
  \begingroup
    \renewcommand{\thefootnote}{\fnsymbol{footnote}}
    \renewcommand{\@makefnmark}{\hbox to \z@{$\^{\@thefnmark}$\hss}}
    \long\def\@makefntext##1{%
      \parindent 1em\noindent
      \hbox to 1.8em{\hss $\m@th \^{\@thefnmark}$}##1
    }
    \thispagestyle{empty}
    \setlength{\tabcolsep}{2pt}
    \vbox{%
      \hsize\textwidth
      \linewidth\hsize
      \vskip 0.1in
      \@toptitlebar
      \centering
      {\LARGE\bf \@title\par}
      \@bottomtitlebar
      \def\And{%
        \end{tabular}\hfil\linebreak[0]\hfil%
        \begin{tabular}[t]{c}\bf\small\rule{\z@}{24\p@}\ignorespaces%
      }
      \begin{tabular}[t]{c}\bf\small\rule{\z@}{24\p@}\@author\end{tabular}%
      \vskip 0.3in \@minus 0.1in
    }
    \@notice
  \endgroup
  \let\maketitle\relax
}
\makeatother

\renewenvironment{abstract}%
{%
  \vskip 0.075in%
  \centerline%
  {\large\bf Abstract}%
  \vspace{0.5ex}%
  \begin{quote}%
}
{
  \par%
  \end{quote}%
  \vskip 1ex%
}

\title{Toward a Compositional Knowledge Architecture for a Shared Organizational Cortex}
\author{
Andrew Miracle$^{\dagger *}$\\
Robin Inc.\\
{\fontsize{8}{9}\selectfont\texttt{me@andrewmiracle.com}}
\And
Phyl Georgiou$^*$\\
Robin Inc.\\
{\fontsize{8}{9}\selectfont\texttt{phyl@withrobin.ai}}
\And
Christine Okoth\\
Robin Inc.\\
{\fontsize{8}{9}\selectfont\texttt{christine@withrobin.ai}}
\And
Ese Monday$^*$\\
Nosen\\
{\fontsize{8}{9}\selectfont\texttt{mondayese8@gmail.com}}
}
\date{}

\begin{document}
\maketitle

\begin{abstract}

Organizations accumulate evidence in conversations, documents, decisions and relationships, but they act on interpretations: judgements assembled from several observations, resting on assumptions, and revised as circumstances change. Existing retrieval, knowledge graphs and agent memory keep the records and lose the interpretation. We propose a four-layer architecture, Signals, Domains, Claims and Dimensions, in which interpretations persist with their justifications and standing, a truth-maintenance network revises that standing as evidence changes, and validated, versioned Dimensions position interpretations on named axes. We call the result a shared organizational cortex. In a registered retrieval experiment, the Dimension term gains 0.028 nDCG@10 (p = 0.002). In system-level comparisons, the full retrieval configuration improves nDCG@10 by 0.081 over Signals-only and 0.053 over graph-augmented Signals-only retrieval (both reported p \textless{} 0.001). The architecture gives organizations a persistent, revisable basis for shared judgement.

\end{abstract}

\section{Introduction}\label{sec:1}

Organizations preserve records, but operational capability depends on judgement: people combine observations from conversations, reports, plans, relationships and experience to assess opportunities, leverage, constraints and assumptions. Such interpretations are provisional [\citelink{cite.ref51}{51}]; theories of change, which state the assumptions through which programs are expected to work [\citelink{cite.ref52}{52}], and investment theses are named forms. Judgement ahead of evidence remains legitimate when its assumptions are explicit.

Search and vector retrieval recover source material, knowledge graphs retain modelled relations, and agent memory retains facts and procedures, but none necessarily preserves what several observations jointly imply. Later requests must then reconstruct the interpretation, possibly differently. Transmission causes the related problem of uncertainty absorption [\citelink{cite.ref34}{34}]: a conclusion travels without the evidence and doubts needed to know whether it still stands.

The requirement is to retain interpretations alongside these systems. Agents increasingly receive per-request context [\citelink{cite.ref46}{46}], while agent memory commonly keeps facts [\citelink{cite.ref09}{9}] or procedures [\citelink{cite.ref31}{31}]. A shared layer should preserve a judgement's provenance, assumptions and latest committed standing while leaving application to the consumer.

Can an organization's accumulated evidence be composed into a persistent, geometrically structured system of interpretations whose justifications, assumptions and standing are maintained as evidence changes, enabling agents to reuse organizational judgment across contexts?

Our organizational cortex composes Entries as source material with four layers: Signals preserve atomic, attributable evidence; Domains provide persistent context; Claims preserve derived interpretations with alternative justifications, evidenced and unevidenced preconditions, provenance and standing; and Dimensions position Claims on named, ordered and interpretable qualities. A dependency network maintains standing as support, conflict and supersession change. Graph edges represent provenance and inferential dependence; geometry represents graded position.

The intended higher-level capability, organizational instinct, is the ability to recognize relevant conditions, reuse justified interpretations, assess positions and constraints, and revise judgement as organizational evidence changes. It is a design objective, neither another layer nor a claim of reproduced human intuition: the cortex is the substrate, instinct the capability it is intended to support.

\begin{figure}[htbp]
\centering
\includegraphics[width=4.35in]{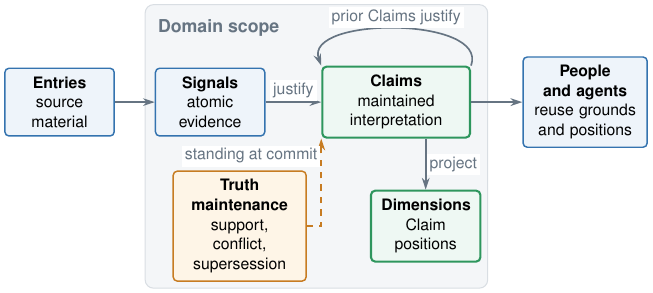}
\caption{The architecture at a glance. Entries supply source material from which multiple Signals preserve atomic evidence and provenance. Domains provide contextual scope. Multiple Signals converge through justification into persistent Claims carrying assumptions and standing; Claims may also depend on prior Claims and are positioned on named Dimensions. Truth maintenance operates over justification, conflict and supersession dependencies at each maintenance cycle. People and agents consume the maintained interpretation with its grounds and geometric positions.}
\label{fig:architecture}
\end{figure}

We contribute (i) the composition of Signals, Domains, Claims and Dimensions into a shared organizational interpretation architecture; (ii) a formal Claim model that retains justifications, assumptions, provenance, standing and revision history; (iii) a truth-maintenance mechanism that propagates support loss without propagating contest; and (iv) a validated, versioned Dimension contract that complements symbolic dependencies with interpretable geometry. We also report (v) a registered retrieval test as evidence about one operational consequence of the representation. We use retrieval to demonstrate whether composed representations affect an observable use of the maintained structure.

\section{From Organizational Experience to Contextual Evidence}\label{sec:2}

Entries are immutable incoming conversations, emails, reports or database exports. From them the system extracts Signals: attributable, embedded propositions that can be judged true or false without the rest of the Entry, references resolved. Because extraction is fallible, each proposition is checked against its source span.

Entries are inputs to the layers. Each Signal belongs to one or more Domains, persistent semantic boundaries such as Climate Change, Kenya or Organizational Relationships that scope Signals, Claims and queries. Their current interpretation, as of the last completed maintenance cycle, is expressed by retrievable Claims at maintained standing.

Persistent objects have stable identity and provenance identifying the relevant source or responsible person or model version. Lifecycle and version changes remain traceable, history is preserved, and temporal provenance distinguishes observation, extraction or derivation, and atomic visibility.

\begin{itemize}
\item \textbf{Entry.} An Entry is immutable source material whose provenance retains its source, receipt time and author where known. Its content is never edited; a person or source system may withdraw it and later restore it. Withdrawal changes the support of Signals extracted from the Entry.
\item \textbf{Signal.} A Signal is an immutable atomic proposition extracted from exactly one Entry and traceable to the source span that entails it by itself. It retains extraction provenance, an embedding for retrieval, one or more Domain memberships, and two distinct times: the observation time stated by the Entry, falling back to the Entry's receipt time, and the later extraction time. It may supersede at most one earlier Signal from the same source when the source marks a replacement or the Labeller finds a restatement with a later value. Removing an incorrect supersession restores the earlier Signal's support. Permitted operations are extraction, classification, reclassification and supersession.
\item \textbf{Domain.} A Domain is a versioned, persistent contextual boundary defined by a name and description, a classification threshold $\theta_D$, and an active Dimension set $A_D$. The Domain classifier asks separately, for each declared Domain description, whether a Signal or query falls within it. A Signal joins $D$ when its returned probability is at least $\theta_D$, or its highest-scoring Domain when none reaches threshold; membership may therefore be multiple. People create, revise, split and retire Domains, with the responsible person, time and derivation history retained. Creation scores every existing Signal against the new Domain; a description revision advances the version and likewise re-scores every Signal against that Domain only. A split creates two Domains derived from the parent and then retires the parent. Reclassification commits atomically and reindexes Claims whose home sets change. Retirement removes the Domain from Signal and Claim home sets, reassigns Signals left without a Domain by the highest-score rule, and parks its Dimensions; split children regain them only through Dimension validation. Changes to $\theta_D$ and $A_D$ remain traceable to the cycle that made them.
\end{itemize}

\begin{figure}[htbp]
\centering
\includegraphics[width=5.50in]{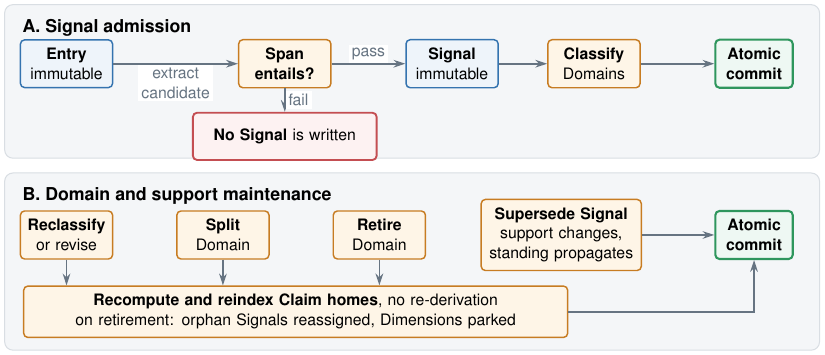}
\caption{Signal admission and Domain maintenance. Only span-entailed candidates become Signals. Supersession changes support; reclassification, revision, splitting and retirement change scope or lifecycle. Affected Claim homes are reindexed, and staged state becomes visible at atomic commit.}
\label{fig:signal-admission}
\end{figure}

Because Domain scope is a hard retrieval filter and a boundary for contradiction candidates, classifier precision and recall matter downstream; the registered classifier test measures them.

{\small
\setlength{\tabcolsep}{3pt}
\setlength{\LTpre}{6pt}
\begin{longtable}{@{}>{\raggedright\arraybackslash}p{0.170\textwidth}>{\raggedright\arraybackslash}p{0.250\textwidth}>{\raggedright\arraybackslash}p{0.290\textwidth}>{\raggedright\arraybackslash}p{0.240\textwidth}@{}}
\multicolumn{4}{@{}p{\textwidth}@{}}{\normalsize Table 1. The four layers and the section that specifies each.}\\[4pt]
\toprule
Layer & Unit & What it holds & Role \\
\midrule
\endfirsthead
\multicolumn{4}{@{}>{\raggedright\arraybackslash}p{\textwidth}@{}}{\normalsize\bfseries Table 1. The four layers and the section that specifies each (continued).}\\[4pt]
\toprule
Layer & Unit & What it holds & Role \\
\midrule
\endhead
\bottomrule
\endlastfoot
Signals & One atomic proposition extracted from one Entry & The proposition, its span and provenance, an embedding, its Domains & The evidence the rest of the system interprets; entailed by its Entry alone \\
Domains & One persistent semantic boundary & A name and description, a classifier threshold, an active set of Dimensions & Semantic routing: classify Signals and scope queries \\
Claims & One proposition no single antecedent entails & Justifications, context, standing, home Domains & Kept interpretations, relabelled as their evidence changes \\
Dimensions & One named, versioned axis & A value space, prototypes, a projection convention, a version & Position Claims and queries for reranking and for the contradiction check's candidate filter \\
\end{longtable}
}

One example runs through the paper. Suppose four Entries yield Signals stating that Foundation X funds climate adaptation programs; Consortium Y operates within several Kenyan counties; the organization has an established relationship with Consortium Y; and county governments control a relevant implementation mechanism. None alone implies a strategy. Asked how the organization could reach Foundation X's county-level climate program, the system retrieves across the Domains in scope and, from the four Signals together, derives: partnering with Consortium Y may provide the strongest route for accessing and implementing Foundation X's county-level climate program. The resulting Claim is homed in every Domain its antecedents contribute and is stored with its justification, so later consumers can reuse the interpretation without deriving it again.

Existing approaches keep different parts of this example: document search ranks the four Entries by lexical overlap with the request, vector retrieval can recover all four Signals, and a knowledge graph can join them by a path query; none of them retains the judgement that the path is the strongest route with a revisable standing. The Claim persists their joint implication and the dependencies through which it changes when Consortium Y leaves the counties, and standing identifies which of ``Foundation X committed \$50 million'' and ``Foundation X cancelled the program'' stands.

Whether a response retrieved or constructed can be tested. A response retrieves when every proposition it presents is a stored object or is entailed by a single stored object, and it constructs an interpretation when it presents a proposition that no single object it cites entails. An interpretation is kept when the system stores it with its justification, and transient otherwise. The partnership proposition is an interpretation by this test, since none of the four Signals entails it. Keeping it is the role of the Claim, whose specification the next section defines.

\section{Claims as Persistent Interpretations}\label{sec:3}

\subsection{The Claim specification}\label{sec:3.1}

A Claim $c$ is a persisted, justified and revisable interpretation. Its immutable natural-language proposition has a stable identity and an embedding used for candidate selection and retrieval; revising the proposition creates a new Claim with a supersedes edge.

The non-empty set $J(c)$ contains alternative justifications. Each $j \in J(c)$ is a non-empty set of antecedent Signals and Claims, with derivation provenance attached to $j$: the model or person, time, and request that produced it or, for corroboration and precondition migration, the write that added it. Unevidenced preconditions remain explicit in the Claim's context. The Claim also retains its current standing, the time-stamped history and causes of standing changes on which the transition rules depend, and any user-rejection or supersession override. Assertion time belongs to derivation provenance; observation times remain properties of the antecedent Signals.

Writing $H(s)$ for a Signal's assigned Domains, the Claim's home set is

\[
H(c) = \bigcup_{j \in J(c)} \bigcup_{a \in j} H(a)
\]

It is recomputed when $J(c)$ or an antecedent's Domain memberships change and is well defined because the justification graph is acyclic. Where the Claim is unambiguous, we write these sets as $J$ and $H$. Derivation, corroboration, precondition migration, relabelling, acceptance, rejection and supersession are permitted operations. Lifted Claims are identified explicitly because their exceptional projection, checking and retrieval behavior differs from ordinary Claims.

The boundary between Signal and Claim is decided by origin and entailment. An object is a Signal if it was extracted from one Entry and its proposition is entailed by that Entry's span alone. An object is a Claim if the system or a user wrote it as an interpretation. A derivation must produce a proposition that no single antecedent entails; one that only restates an antecedent writes nothing and cites it instead. Both entailment conditions are checked by the Labeller, once per extracted Signal and per derivation antecedent; a misjudged object can surface through the contradiction check or a user's rejection. A supports verdict from the contradiction check is corroboration, not derivation, so the rule does not apply to it. Two exceptions are restatements by construction: the lifted Claim, never retrievable, and a user-entered Claim, justified by the Signal extracted from the Entry recording the user's statement plus any objects the user cites.

\subsection{Where Claims live, and what context holds}\label{sec:3.2}

A Claim is stored once and indexed under every Domain in $H$. Its preconditions cannot remain implicit because later requests may use it under different assumptions.

An evidenced precondition is one that some Signal or Claim in the workspace asserts. When a Claim is derived under an assumption, the system must search for a supported Signal or Claim asserting it; one that is found joins every justification of the new Claim, so the assumption lives in $J$ and the network retracts the Claim when that evidence is withdrawn or superseded (Section 6 follows one such precondition).

An unevidenced precondition is one for which no evidence was found; only these are stored in context. Any downstream presentation must expose each beside the Claim it uses, and every derivation or revision that touches the Claim repeats the search. A precondition that acquires evidence moves into $J$.

\section{Maintaining Organizational Judgment}\label{sec:4}

Truth maintenance is the mechanism that makes persistence safe under change. Without it, a repository can retain an interpretation after its evidentiary basis has disappeared. The mechanism operates over Claims and their dependencies.

\subsection{Standing}\label{sec:4.1}

Standing is the Claim's status in the system's current committed view of the evidence, one of five states: proposed, accepted, contested, retracted, and superseded. Throughout this paper, ``current'' standing or interpretation means current as of the last completed maintenance cycle. The states and their transitions are:

\begin{itemize}
\item \textbf{proposed}: the initial state of any Claim written by the system or entered by a user, unless it arrives with no supported antecedent and is retracted at once. A proposed Claim is retrievable and penalized by the standing term $\mu$ while the registered retrieval test keeps it.
\item \textbf{proposed to accepted}: a user accepts the Claim, or the network applies the automatic acceptance rule: at least one justification has antecedent Signals from at least two distinct Entries, all of that justification's antecedents are supported, the contradiction check found no conflicting Claim, and the Claim has never been retracted. The rule is evaluated when the check completes and whenever propagation re-evaluates the Claim. Because contest does not travel along justifications, a Claim resting on a contested Claim can still be accepted. Workspaces may disable the automatic rule and require a person for every acceptance.
\item \textbf{accepted or proposed to contested}: the dependency network records a conflicts\_with edge between this Claim and another Claim whose standing is neither retracted nor superseded. Contested Claims remain retrievable, penalized by the reranker, and any consumer using one must be shown the conflict.
\item \textbf{contested to accepted, or contested to proposed}: the conflicting partner becomes retracted or superseded (by propagation, by a user rejecting it, or by a supersedes edge), or a user marks the conflict spurious, deleting the edge. The Claim returns to accepted if it had been accepted before the conflict and to proposed otherwise.
\item \textbf{any state to retracted}: no justification remains supported, or a user rejects the Claim. Retracted Claims are excluded from retrieval and kept for audit.
\item \textbf{retracted to proposed}: a Claim retracted by propagation regains support because a withdrawn Entry is restored or a superseding Signal is itself withdrawn; since the automatic rule excludes Claims ever retracted, only a person can accept it again. A Claim retracted by a user does not re-enter on regained support.
\item \textbf{any state to superseded}: the system or a user writes a revised Claim with a supersedes edge to this one. The superseded Claim is excluded from retrieval by default and remains reachable from its successor; deleting a wrong edge returns it to the standing its support and history give.
\end{itemize}

\begin{figure}[htbp]
\centering
\includegraphics[width=4.47in]{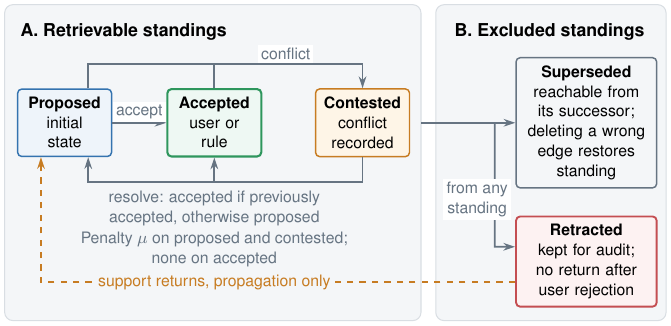}
\caption{Claims begin proposed, become accepted by review or rule, become contested under live conflict, retract on support loss or rejection, and become superseded on revision; a propagated retraction restored to support returns proposed. Panel A contains retrievable proposed, accepted and contested Claims, with penalties on proposed and contested; panel B contains retracted Claims kept for audit and superseded Claims reachable from successors.}
\label{fig:standings}
\end{figure}

\subsection{Justification dependencies and propagation}\label{sec:4.2}

The system maintains standing with a justification-based dependency network, adapting the justification-based truth maintenance approach of Doyle [\citelink{cite.ref15}{15}].

Nodes are Signals and Claims. A Signal is supported unless its Entry is withdrawn or a later same-source Signal with a live Entry supersedes it. A Claim is supported when it has no override and some non-empty justification has every antecedent supported. A symmetric conflicts\_with edge records which contradiction-check branch asserted it. Supersession and user rejection override support, setting standing to superseded and retracted respectively, with rejection taking precedence; an override persists if a successor later loses support, and deleting an erroneous supersedes edge removes it. Propagation neither changes overrides nor treats overridden Claims as supported, so an override propagates to dependents. Deleting a conflict edge is an edit, not an override. Without an override, an unsupported Claim is retracted; a supported Claim with a live conflict partner is contested; otherwise it is accepted if previously accepted and not since retracted, and proposed if not.

\begin{figure}[htbp]
\centering
\includegraphics[width=4.00in]{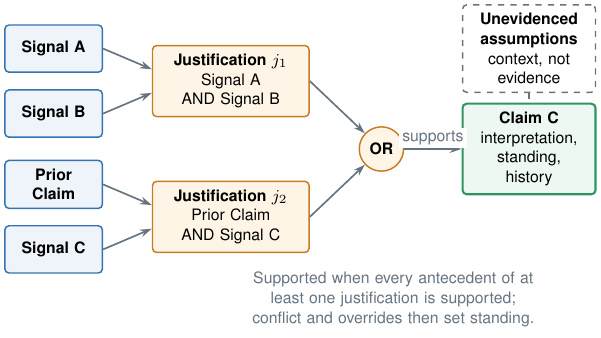}
\caption{Claim support structure. Within justification $j_1$, Signal A and Signal B are jointly required (AND); $j_2$, supported by a prior Claim and Signal C, is an alternative support path (OR). Unevidenced assumptions remain in context outside the evidence dependency graph. Claim C retains its interpretation, standing and history. The Claim is supported when every antecedent in at least one justification is supported; conflict and overrides then determine standing.}
\label{fig:claim-support}
\end{figure}

When support changes, the network collects the changed node's forward justification closure, re-evaluates each Claim once in topological order, then once re-evaluates the conflict partners of closure Claims, including the changed node, whose standing changed. A partner's support is unchanged, so it propagates no further. Signal support has one dependence on absence, a live superseder, but supersession and Entry withdrawal do not depend on labels; Signals therefore form a uniquely labelled, well-founded bottom stratum. Above it, Claim justifications are monotone and acyclic, so one pass produces the unique well-founded support labelling regardless of arrival order. Standing may remain order-dependent because it reads history. Doyle's non-monotonic justifications are excluded: ``no competing bid has been submitted'' requires a Signal asserting it.

Support is also well-founded [\citelink{cite.ref18}{18}]: mutually justifying Claims with no Signal beneath them are unsupported. Derivation is acyclic because antecedents pre-exist the new Claim. Before a supports verdict, re-derivation or precondition migration adds an antecedent to an existing Claim, a reachability search linear in that antecedent's ancestors rejects the write if the Claim is its ancestor.

Relabelling updates standing over the closure and its conflict partners, visiting each node at most once; even when the closure spans the workspace, maintenance remains a deterministic graph traversal. Contest stops at the two Claims joined by conflicts\_with.

When a recently retrieved Claim becomes retracted or contested, re-derivation reruns its question on present evidence: a different proposition supersedes it, the same proposition on new evidence adds a justification, and otherwise nothing is written. Retrieval marks limit this to Claims used within thirty days and one derivation per relabelled Claim, so the worst case remains the closure and its partners.

Changes to a Domain are processed in atomic batches, one per maintenance cycle, flushed at an interval $\Delta$ ($\Delta = 5$ minutes initially). Within a batch, a new Claim is projected onto at most $K$ active Dimensions of its home Domains before its contradiction check, so the check sees positions; that requires at most $K$ projections per Claim, by the reference projector until another model qualifies.

The contradiction check runs when a Signal or Claim is written to a Domain. For a Claim, candidates are its $k$ nearest Claims sharing a home Domain by embedding similarity ($k=20$ initially), restricted to a shared Dimension $d$ with relevance $r_d\geq\tau_r$ for both and distance $\delta_d>\tau_\delta$ ($\tau_r=\tau_\delta=0.5$ initially). Fewer than five survivors triggers similarity-only fallback, so missing Dimension coverage grants no exemption. For an unprojected Signal, candidates are the $k$ nearest Claims above $\tau_s$, the Domain's lowest decile of Signal-to-nearest-Claim similarities. The Labeller, by default an NLI model [\citelink{cite.ref05}{5}, \citelink{cite.ref53}{53}], labels each pair supports, contradicts or unrelated. Contradiction writes a branch-tagged conflicts\_with edge; support adds a single-antecedent justification subject to the cycle check. Adding justification can only add support; precondition migration alone narrows it by making an assumed dependence explicit.

Three kinds of conflict are excluded. Signals are not checked against Signals, since disagreement is recorded only where an interpretation rests on it. A Claim written after a Signal that contradicts it is checked only against Claims, the Deriver's reading of the Domain's Signals being the only guard in that direction. And objects sharing no Domain are never compared, since the check runs within a Domain, so a conflict across disjoint Domains surfaces only if a later derivation spans both.

Because conflicts\_with joins two Claims, a contradicting Signal is first lifted into a proposed Claim that restates its proposition and is justified by that Signal alone, and the edge joins the lifted Claim to the one it contradicts. The lifted Claim is supported while its Signal is; it is not projected, triggers no check and is not retrievable. A later ordinary Claim with the same proposition takes over the edge. Candidate filtering bounds the check at $k$ classifier calls plus one lifting call per Signal write.

If a different source reports that Consortium Y withdrew from the counties, a contradicts verdict lifts the Signal, writes the edge and contests the accepted partnership Claim. Section 6 follows the same-source case, where supersession retracts it.

\section{Dimensions and Geometric Knowledge Representation}\label{sec:5}

\subsection{Explicit Graphs and Geometric Relationships}\label{sec:5.1}

Graph edges represent named relations such as extraction, authorship, classification, justification, citation, conflict and supersession. Dimensions represent opposite positions and betweenness on an axis. Gärdenfors' conceptual spaces place entities on interpretable quality dimensions and relate similarity to distance [\citelink{cite.ref21}{21}]; named axes identify where Claims are close or far.

Gärdenfors [\citelink{cite.ref20}{20}] calls an integral set of quality dimensions, such as hue, saturation and brightness, a ``domain.'' Here Domain means an organizational context that scopes Signals and reads; we call his usage a quality domain.

A Dimension is an axis on which knowledge objects can meaningfully be compared, such as ``funding availability''.

A specified axis is auditable: every coordinate has a definition, a projector and a version, so a ranking is explained by naming the axis and checked by re-projecting. It is stable: a Dimension changes one axis at a time, under a version.

\subsection{The Dimension contract}\label{sec:5.2}

A Dimension's contract has seven fields,

\[
d = (V_d,\ \Pi_d,\ \rho_d,\ \nu_d,\ P_d,\ \text{prov}_d,\ \text{ver}_d)
\]

each drawn from or checked against a controlled registry:

\begin{itemize}
\item $V_d$, the value space and its range: a bounded scalar in $[0, 1]$ or an ordered scale of named levels. Unordered categories are not admitted, since an axis whose values have no order is a facet, not a Dimension.
\item $\Pi_d$, the projection system: a projector model and a prompt filled with the Dimension's name, definition text and prototypes, or a trained head, returning a value in $V_d$ and a relevance in $[0, 1]$.
\item $\rho_d$, the comparison function on $V_d$: absolute difference for scalars, rank distance for ordered scales.
\item $\nu_d$, the normalizer that turns $\rho_d$ into $\delta_d \in [0, 1]$, for instance division by the range or by the number of levels minus one.
\item $P_d$, two to five prototypes: reference propositions with assigned values, shown to the projector as anchors. They calibrate projection by anchoring the projector.
\item $\text{prov}_d$: who proposed it (model proposal or disentanglement with the model identifier, or a named person for a human-authored Dimension), when, the Claim-pair sample it was proposed from, its validation results with the projectors used, and what activated it, the Useful screen or a named user's pin.
\item $\text{ver}_d$, an integer version. Outside the version, each Dimension also has an identifier, an owning Domain and a lifecycle status (candidate, validated, active, parked or retired); its name and definition text are part of $\Pi_d$.
\end{itemize}

Implementation feasibility, used below, sets $V_d$ to (blocked, constrained, feasible, ready), uses rank distance and normalizes by three. An interpretable axis has a name and definition, a stated projection procedure, and assignments checked by Measurable and Stable.

The contract retains per-Dimension metric structure and separability: weighted terms are city-block in form [\citelink{cite.ref20}{20}]. Per-Dimension distances add, and each term is also scaled by Claim relevance, so the sum is a relevance-weighted similarity rather than a metric. It retains single-axis betweenness, where position $y$ lies between positions $x$ and $z$ when $\delta_d(x,y)+\delta_d(y,z)=\delta_d(x,z)$. It defines no integral grouping, cross-Dimension betweenness, convex concept regions or prototype categorization. Models propose and validate Dimensions, and people handle replacements, pins and retirement.

\subsection{A constrained Dimension specification}\label{sec:5.3}

Models compose approved value spaces, projection systems, metrics and normalizers from a controlled registry; a typed schema rejects mismatches such as rank distance on a scalar. The normalizer maps $\rho_d$ to $\delta_d(x, y) \in [0, 1]$ for positions $x, y \in V_d$, with $0$ for indistinguishable and $1$ for maximally different positions, allowing comparison across native representations.

\subsection{Position, relevance, distance and salience}\label{sec:5.4}

For Claims $c, c'$ and Dimension $d$, position $\pi_d(c)\in V_d$ says where $c$ lies; relevance $r_d(c)\in[0,1]$ says how strongly it concerns $d$; distance $\delta_d(\pi_d(c),\pi_d(c'))$ compares positions; and fitted salience $\beta_{D,d}$ says how much $d$ matters in Domain $D$. In retrieval, $r_d(c)$ enters $\phi_d(c, q)$ as a continuous factor and is not thresholded. The threshold $\tau_r=0.5$ applies to Covered, contradiction filtering and query position, and $\tau_g$ gates projection by embedding; Section 8's graded judgments are separate human labels. Pilot applicability alpha ranged from $-0.37$ to $0.20$, with one pair showing $33\%$ versus $97\%$ coverage (Table 5), leaving the relevance cut and Covered floor least settled. Covered is the projector's relevant share. Applicability is measured separately from position.

The commitment and cancellation in Section 2 are both relevant to funding availability and close in embedding space but far apart on that axis, the signature used by the Claim contradiction branch.

Projection returns $(\pi_d(c),r_d(c))$ for one Claim and Dimension; Signals are never projected. One-Dimension calls keep validation agreement unconfounded (Table 3). Fine-tuned models can match larger models at recovering conceptual-space rankings [\citelink{cite.ref07}{7}], so a reference projector may bootstrap labels for a trained replacement [\citelink{cite.ref50}{50}] that qualifies through the Measurable, Stable and registered Projector tests; qualification is the criterion.

\subsection{Discovery and validation}\label{sec:5.5}

Candidate Dimensions are validated, activated if useful, and eventually parked or retired; $A_D$ denotes Domain $D$'s active set. They have two non-exclusive sources. Each evaluation cycle (initially seven days) proposes zero or more constrained specifications from the Domain description, active definitions and embedding-close pairs the reranker separates. A second process names directions disentangled from Domain Claim embeddings. Dimensions belong to Domains because interpretable features often apply only to subsets of entities [\citelink{cite.ref03}{3}].

\begin{figure}[htbp]
\centering
\includegraphics[width=5.50in]{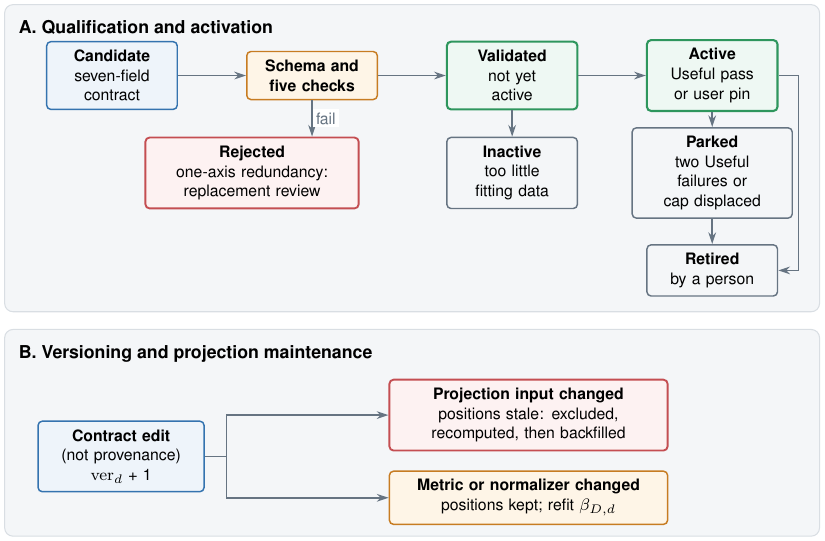}
\caption{Dimension qualification, activation and projection lifecycle. A candidate supplies the seven-field contract and passes schema and qualification checks before validation. Useful or pin activation, inactive validation, replacement review, parking and human retirement remain distinct. Version edits either stale projections for exclusion and backfill or retain values while coefficients are refit.}
\label{fig:dimension-lifecycle}
\end{figure}

Validation uses a Domain sample $S$ ($n = 100$ where available; minimum $30$). Table 2 gives the operational tests and tunable initial thresholds; $r_s$ denotes Spearman agreement, distinct from $\rho_d$.

{\small
\setlength{\tabcolsep}{3pt}
\setlength{\LTpre}{6pt}
\begin{longtable}{@{}>{\raggedright\arraybackslash}p{0.170\textwidth}>{\raggedright\arraybackslash}p{0.250\textwidth}>{\raggedright\arraybackslash}p{0.290\textwidth}>{\raggedright\arraybackslash}p{0.240\textwidth}@{}}
\multicolumn{4}{@{}p{\textwidth}@{}}{\normalsize Table 2. Dimension qualification matrix. The criteria test distinct properties of a candidate Dimension and its projection procedure. Passing them establishes operational consistency under the stated tests.}\\[4pt]
\toprule
Criterion & Qualification procedure & Initial pass bar or timing & Caveat or consequence \\
\midrule
\endfirsthead
\multicolumn{4}{@{}>{\raggedright\arraybackslash}p{\textwidth}@{}}{\normalsize\bfseries Table 2. Dimension qualification matrix (continued).}\\[4pt]
\toprule
Criterion & Qualification procedure & Initial pass bar or timing & Caveat or consequence \\
\midrule
\endhead
\bottomrule
\endlastfoot
Measurable & Project $S$ independently using different prompt phrasings or the two projector models named under Stable & $r_s \geq 0.7$ & Tests rank agreement; the interval spacing assumed by $\delta_d$ and betweenness is untested. \\
Covered & Measure the share of $S$ with relevance $r_d \geq \tau_r$ & At least $20\%$ & Below the floor the candidate fails qualification. \\
Distinct & Compare the candidate definition embedding with every existing Dimension definition in the Domain & Cosine similarity $< 0.9$ & Prevents near-duplicate definitions \\
Non-redundant & Correlate positions with each active Dimension on Claims relevant to both & $|r_s(\pi_d, \pi_{d'})| < 0.8$ & Failure against exactly one axis creates a replacement candidate for human review. \\
Stable & Re-project with a different model at validation and after time or corpus growth; repeat after any version bump that stales projections & Meet Measurable at validation and again after 30 days or $20\%$ Claim-count growth, whichever comes first & Tests agreement across projector and time changes \\
\end{longtable}
}

These tests establish consistency across prompts, projectors and time; Section 8.3 reports applicability.

A validated candidate is projected only for Claims retrieved by validation queries; activation triggers Domain-wide projection. The Useful screen refits the reranker by five-fold cross-validation and requires both non-zero $\beta_{D,d}$ and higher out-of-fold nDCG@10 in a majority of folds; ties fail. The Useful screen runs on validation data; the registered test uses held-out queries. Without enough fitting data the Dimension stays inactive; a user pin activates it regardless, two consecutive screen failures park it (kept but not projected or used), and only a person retires it.

The apparent cycle between discovery and standing is temporally stratified: discovery reads text, embeddings and reranker residuals fixed at cycle start, never standing; standing reads positions only through a non-writing candidate filter; and new Dimensions affect only later checks. Active sets are fixed per batch.

\subsection{Projection lifecycle and invalidation}\label{sec:5.6}

Three controls bound projection: Claim--definition cosine similarity $\hat r_d(c)$ must clear permissive gate $\tau_g$ (initially the Domain's 30th percentile); each Domain has $|A_D|\leq K$ ($K=8$ initially), with a newly Useful Dimension displacing the smallest mean out-of-fold $\beta_{D,d}$ and ties parking the most recent; and a multi-Domain Claim receives at most $K$ projections, selected by largest mean $\beta_{D,d}$. Projection occurs when a Claim is written or a Dimension activated, before contradiction checking, never during ordinary retrieval.

Each projection retains Dimension/version, position, relevance, provenance and time; compatibility rules derive validity. Any contract change except provenance increments $\text{ver}_d$. Changes to projection inputs (value space, system or head, definition or prototypes) stale earlier positions; metric or normalizer changes preserve positions but refit $\beta_{D,d}$ next cycle. Stale projections are excluded from reranking and contradiction filtering; retrieved Claims queue for next-batch re-projection and the rest backfill by access frequency. Retired projections remain only for audit. A version bump can thus invalidate every projection in the Domain, motivating Stable.

\subsection{Domain scope and Dimension-aware retrieval}\label{sec:5.7}

The architecture routes queries and Signals to Domains before positioning Claims on those Domains' Dimensions.

For a candidate Claim $c$ retrieved for query $q$, with hybrid base score $b(c, q)$ from the fusion-and-cross-encoder stage, define the per-Dimension feature

\[
\phi_d(c, q) = r_d(c) \cdot \big(1 - \delta_d(\pi_d(c), \pi_d(q))\big)
\]

where a qualified Projector obtains query position $\pi_d(q)$ under the one-Dimension convention, using at most $|A_D|$ parallel calls or one encoder pass with independent heads, never the Deriver. If $r_d(q)<\tau_r$, the query expresses no position: replacing $(1-\delta_d)$ by $1$ reduces $\phi_d$ to Claim relevance. Our post hoc facet-only arm in Section 8.5 isolates this component from positional geometry. The score is

\[
\text{score}(c, q) = b(c, q) + \sum_{d \in A(c)} w_q(d)\, \phi_d(c, q) - \mu \cdot [\text{standing}(c) \in \{\text{contested},\ \text{proposed}\}]
\]

where $A(c)$ is the set of at most $K$ Dimensions, drawn from the active sets of $c$'s home Domains, on which $c$ is projected and has a non-stale projection, and $[\cdot] \in \{0, 1\}$ equals $1$ when its condition holds and $0$ otherwise.

The Dimension term reads positions and fitted coefficients.

\begin{figure}[htbp]
\centering
\includegraphics[width=5.50in]{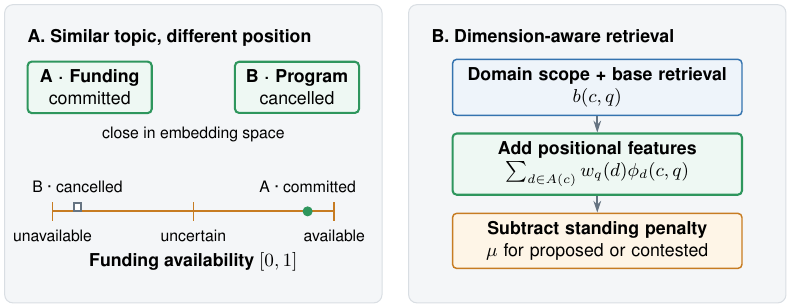}
\caption{Complementary relations. Panel A contrasts embedding proximity with separated funding-availability positions. Panel B adds positional Dimension features to Domain-scoped base retrieval and subtracts the proposed-or-contested standing penalty.}
\label{fig:geometric-relations}
\end{figure}

For $D$ and $d\in A_D$, pairwise learning-to-rank [\citelink{cite.ref06}{6}] fits $\beta_{D,d}$ over $\phi_d$ and graded validation labels, constrained non-negative and $L_1$-penalized [\citelink{cite.ref48}{48}]. The weight $w_q(d)$ is $\beta_{D,d}$ when $d$'s owning Domain $D$ is in $q$'s scope and $0$ otherwise; we fit $\mu \geq 0$ jointly or alone as in the registered test. Fewer than 30 in-scope validation queries leaves Dimensions inactive and the Domain unfit. Coefficients refit each evaluation cycle, remain constant between cycles and take effect at batch commit.

\section{Composition into a Shared Organizational Cortex}\label{sec:6}

Signals make sources attributable; Domains establish context; Claims retain interpretations and grounds; Dimensions enable ordered comparison; and truth maintenance revises standing over justification, conflict and supersession. Claims retain interpretation, maintenance keeps its standing current, justification dependencies explain why it stands, and geometry gives it graded position.

The reusable unit is therefore the Claim plus its justification graph and preconditions, latest committed standing, and current versioned Dimension projections. These explain why it was formed and what can withdraw support, whether it may be reused or carries conflict, and where it lies on relevant qualities. The proposition is fixed; standing records support, position records location on a quality, and justification records derivation.

By preserving interpretations together with their justifications, assumptions, standing and positions, the architecture enables subsequent consumers to reuse organizational judgment without reconstructing its supporting relationships from source evidence. Through any home Domain, a consumer can recover the same Claim, inspect its grounds and committed standing, and judge how it applies to the present task. The cortex supports organizational instinct by making prior judgement accountable, positioned and revisable.

Retrieval scopes Domain-assigned Signals and Claims, combines BM25 [\citelink{cite.ref43}{43}] and dense retrieval [\citelink{cite.ref28}{28}], applies cross-encoder [\citelink{cite.ref33}{33}, \citelink{cite.ref37}{37}] and Dimension-aware reranking over stored provenance, standing and positions, and returns cited maintained objects. A broader reasoning harness may consume the same Claims and grounds for subsequent judgement.

Deferred maintenance over a materialized view [\citelink{cite.ref11}{11}] commits accumulated changes atomically after $\Delta$, so consumers never see unchecked Claims or partial propagation. Staleness is at most one version batch but $\Delta$ plus unbounded processing time; overload delays later batches. Session reads are monotonic, and the write operation sees its Claim, although pre-commit retrieval does not [\citelink{cite.ref47}{47}]. Content staleness also depends on contradiction recall and recent-use-limited re-derivation, which our registered replay series tests. Retrieval records marks and queues stale projections.

Table 3 fixes bounded outputs independently of model family. Qualification binds an exact model version; replacements, including references, rerun the registered test, and failure drops the dependent component as specified. Retrieval embeddings, cross-encoders and pre-check Extractor proposals sit outside the qualified role set; Deriver and Labeller never run during ordinary retrieval.

{\small
\setlength{\tabcolsep}{3pt}
\setlength{\LTpre}{6pt}
\begin{longtable}{@{}>{\raggedright\arraybackslash}p{0.160\textwidth}>{\raggedright\arraybackslash}p{0.190\textwidth}>{\raggedright\arraybackslash}p{0.240\textwidth}>{\raggedright\arraybackslash}p{0.170\textwidth}>{\raggedright\arraybackslash}p{0.180\textwidth}@{}}
\multicolumn{5}{@{}p{\textwidth}@{}}{\normalsize Table 3. Essential model responsibilities and output constraints.}\\[4pt]
\toprule
Name & Responsibility & Input & Permitted output & Constraint \\
\midrule
\endfirsthead
\multicolumn{5}{@{}>{\raggedright\arraybackslash}p{\textwidth}@{}}{\normalsize\bfseries Table 3. Essential model responsibilities and output constraints (continued).}\\[4pt]
\toprule
Name & Responsibility & Input & Permitted output & Constraint \\
\midrule
\endhead
\bottomrule
\endlastfoot
Extractor & Propose atomic evidence from an Entry & One Entry & Candidate propositions with source spans & A proposal is stored only after the Labeller confirms span entailment; Extractor recall is not tested \\
Deriver & Compose an interpretation from several antecedents & A request with retrieved Signals and Claims & A Claim proposition with justifications and context & It cannot set standing \\
Labeller & Judge entailment, support or contradiction & A span or object with a candidate proposition & Entailed or not; supports, contradicts or unrelated & It cannot write propositions \\
Projector & Position a Claim or query on one Dimension & One proposition or query and one $\Pi_d$ & One value in $V_d$ and one relevance & It cannot produce prose or take several Dimensions per call \\
Domain classifier & Assign contextual scope & One Signal or query and declared Domain descriptions & One probability per declared Domain & It cannot name an undeclared Domain \\
\end{longtable}
}

A general-purpose or fine-tuned model, head or adapter [\citelink{cite.ref26}{26}] may qualify. A trained projector head belongs to $\Pi_d$, so retraining increments $\text{ver}_d$.

A persisted Claim carries durable justifications, withdrawal-sensitive standing and cross-time conflicts, because maintenance stays attached.

The running example shows the composition, with illustrative values:

\begin{itemize}
\item \textbf{Entry.} A partner profile arrives from the relationship records at time $t_0$, authored by the partner manager.
\item \textbf{Signal.} Extraction yields ``Consortium Y operates within several Kenyan counties'', retaining the source span, extraction provenance and time, and an observation time taken from the profile's date.
\item \textbf{Domains.} The classifier scores it above threshold for Kenya and Organizational Relationships, and it is indexed in both.
\item \textbf{Claim.} A request about Foundation X's county program derives the partnership Claim of Section 2 from the four Signals, this one among them, plus a Signal found for the precondition that the program is still open. $H$ is the union of the antecedents' Domains; standing is proposed.
\item \textbf{Dimension.} In the same batch the Claim is projected on implementation feasibility; suppose the projector returns feasible with relevance $0.9$, recorded under the Dimension's current version. The check finds no conflict, the antecedents come from distinct Entries, and the automatic rule accepts the Claim. The batch commits.
\item \textbf{Reuse.} A later agent asks which partners can implement in Kenyan counties. Suppose the query projects to ready, the top level, with relevance $0.8$: $\delta_d = 1/3$, so $\phi_d = 0.9 \cdot (1 - 1/3) = 0.6$ enters the score with weight $\beta_{D,d}$, and the Claim is returned with its citations, assumptions and standing without reconstructing the interpretation.
\item \textbf{Change.} The same source later supersedes the Signal. Propagation retracts the partnership Claim, whose only justification has lost an antecedent, and every Claim justified only through it. After the batch containing the supersession and propagation commits, the next read no longer returns the Claim; because it was recently retrieved, re-derivation is scheduled and writes a superseding Claim if the remaining evidence supports another route.
\end{itemize}

\begin{figure}[htbp]
\centering
\includegraphics[width=4.45in]{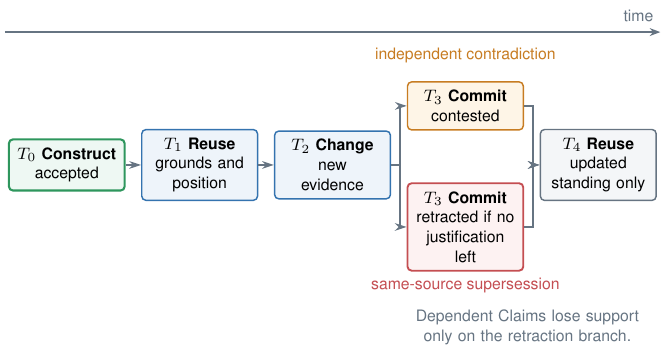}
\caption{Foundation X lifecycle. Signals justify and position the Claim at $T_0$; an agent reuses its grounds, assumptions, accepted standing and position at $T_1$. At $T_2$, independent contradiction creates conflict while supersession removes support. The $T_3$ commit contests the first branch and retracts the second if no justification survives; only retraction propagates support loss. At $T_4$, consumers receive the updated standing.}
\label{fig:foundation-lifecycle}
\end{figure}

Across both branches, the proposition and its provenance remain in history; maintenance changes committed standing, and reuse follows the current standing.

\section{Related Work and Discussion}\label{sec:7}

Gärdenfors [\citelink{cite.ref21}{21}] supplies the representational level occupied by Dimensions. Bechberger and Kühnberger [\citelink{cite.ref04}{4}] formalize metric betweenness and replace convex regions, which under city-block distance cannot represent correlations between domains, with fuzzy star-shaped sets; Section 5.2 states the narrower properties retained here. At a different level, proposition units outperform passages for retrieval and passage-limited question answering [\citelink{cite.ref08}{8}], supporting Signals but also motivating the Signals-only arm that tests whether Claims add value. RAPTOR precomputes a tree of clustered chunk summaries [\citelink{cite.ref44}{44}] but does not maintain derived interpretations as evidence changes.

GraphRAG extracts entities, relations and source-linked claims and pregenerates community summaries [\citelink{cite.ref17}{17}]; LightRAG updates such a graph incrementally [\citelink{cite.ref22}{22}]. Neither retains the justification sets and standing defined here, although a GraphRAG-style index underlies the graph arms. Zep's Graphiti temporal graph timestamps facts and invalidates overlapping contradictions [\citelink{cite.ref42}{42}], making it the nearest deployed comparison and the basis of a registered replay arm. Contradiction remains difficult: de Marneffe, Rafferty, and Manning [\citelink{cite.ref13}{13}] reported about 23\% precision and 19\% recall and prioritized precision because contradictions are rare; despite stronger modern NLI benchmarks [\citelink{cite.ref24}{24}], the standing test therefore measures precision and recall separately against a reference Labeller.

Doyle's TMS [\citelink{cite.ref15}{15}] stores belief justifications and revises status; de Kleer's ATMS [\citelink{cite.ref12}{12}] labels the assumption sets under which nodes hold. In Section 4 we use a well-founded, otherwise monotone subset of Doyle's network and do not adopt ATMS contexts for unevidenced preconditions. Incremental view maintenance counts derivations or deletes and rederives recursive views [\citelink{cite.ref23}{23}]; acyclicity permits one topological pass here. AGM [\citelink{cite.ref02}{2}] addresses logically closed theories, unlike the Claim set, but minimal change corresponds to retracting exactly Claims whose every justification fails. Dung's attack relations and extension semantics [\citelink{cite.ref16}{16}] frame conflicts\_with; we deliberately avoid computing an extension for contested status, leaving grounded semantics for future conflict chains.

PROV-DM's entities, activities and agents [\citelink{cite.ref36}{36}] map to Entries, Signals and Claims; extraction and derivation; and their models or people. Justifications and Domain lineage are derivations, supersession a revision and withdrawal an invalidation. Domains scope context, Dimensions position Claims on axes, and graph edges record provenance and dependence. Noy and Klein's multidimensional account of ontology-version compatibility [\citelink{cite.ref39}{39}] motivates separating Dimension changes that invalidate positions from those altering only comparisons.

Organizational knowledge creation spans tacit and explicit knowledge [\citelink{cite.ref38}{38}]; our design addresses only the explicit side. Knowledge-management repositories support creation, transfer and application but ordinarily store curated documents without support or standing [\citelink{cite.ref01}{1}]. Generative agents retain reflections with memory pointers but do not revise them when those memories are contradicted [\citelink{cite.ref40}{40}], while Zep revises facts rather than propositions derived from them. BeliefBank [\citelink{cite.ref29}{29}] and REFLEX [\citelink{cite.ref30}{30}] restore one model's consistency using MaxSAT or constraint reasoning; here persisted propositions are relabelled through evidence-backed justifications.

Longer contexts expose more evidence but neither preserve a shared interpretation nor check it when evidence changes; request-keyed caches neither transfer it across phrasings nor revise it. Maintenance also imposes curation and may not benefit every workload. Our contribution is the composition of known parts: truth maintenance over derived propositions supported by evidence and prior Claims; validated, versioned axes used for comparison and contradiction candidates; and their integration with Signals and Domains into a reusable substrate with grounds, standing and positions.

{\small
\setlength{\tabcolsep}{3pt}
\setlength{\LTpre}{6pt}
\begin{longtable}{@{}>{\raggedright\arraybackslash}p{0.160\textwidth}>{\raggedright\arraybackslash}p{0.190\textwidth}>{\raggedright\arraybackslash}p{0.240\textwidth}>{\raggedright\arraybackslash}p{0.170\textwidth}>{\raggedright\arraybackslash}p{0.180\textwidth}@{}}
\multicolumn{5}{@{}p{\textwidth}@{}}{\normalsize Table 4. The systems this section compares, on the properties of our design claims.}\\[4pt]
\toprule
System & Knowledge unit & Invalidation and standing & Multi-axis reranking & Persistence and update model \\
\midrule
\endfirsthead
\multicolumn{5}{@{}>{\raggedright\arraybackslash}p{\textwidth}@{}}{\normalsize\bfseries Table 4. Systems compared (continued).}\\[4pt]
\toprule
System & Knowledge unit & Invalidation and standing & Multi-axis reranking & Persistence and update model \\
\midrule
\endhead
\bottomrule
\endlastfoot
Retrieval with per-request reasoning (the alternative) & Retrieved Signals and documents, nothing derived is kept & None: there is no persisted statement to invalidate & The base retrieval score only & Interpretations are derived per request and discarded \\
GraphRAG and LightRAG & An entity graph; claims extracted from text units; community summaries & Extracted claims are produced once with a status fixed at extraction, and are not revised when the source changes & No Dimension term & Batch indexing; summaries pregenerated for global questions \\
Zep (Graphiti) & Temporal fact edges in a knowledge graph & Edges are invalidated on overlapping contradictions & No Dimension term & The graph is updated incrementally as facts arrive \\
This architecture & Signals, and Claims with justifications over them & Standing is relabelled by propagation through justifications: conflict contests, support loss retracts, a revision supersedes & The Dimension term over validated, versioned Dimensions, plus the standing penalty & Interpretations persist and are maintained as evidence changes \\
\end{longtable}
}

\section{Empirical Evaluation of Representational Utility}\label{sec:8}

The evaluation tests representational utility; Figure 8 maps evidentiary status.

\begin{figure}[htbp]
\centering
\includegraphics[width=4.30in]{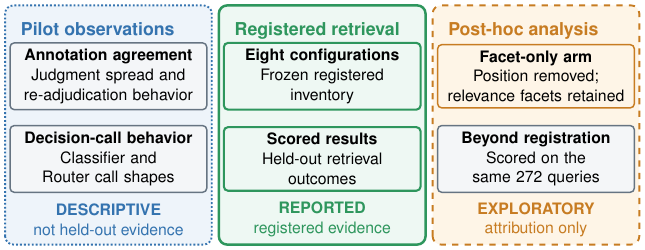}
\caption{Evaluation scope and evidentiary status. Pilot observations characterize decision-call behavior and annotation variability. The registered retrieval experiment provides the reported empirical results, supplemented by a post hoc facet-only comparison. The remaining registered components are specified in the evaluation protocol but fall outside the empirical scope of this paper.}
\label{fig:evaluation-status}
\end{figure}

\subsection{Setup}\label{sec:8.1}

\textbf{Domains.} From public sources, the pilot kit built 121 Signals from 49 packaging Python Enhancement Proposals and 56 from 239 messages in one Enron custodian's mailbox. Each Domain had 30 Claims and 6 queries with 6 candidates, hence 36 graded pairs. An AI coding assistant prepared these fixed, non-system inputs and checked every Signal span verbatim.

\textbf{Annotators.} A five-item calibration preceded annotation, with one non-blind calibration excluded.

\textbf{Decision calls.} Models clef and jev classified the queries by forced two-way choice and by per-Domain yes/no calls, one per item returning a probability per Domain; the latter also covered all 177 Signals. Appendix A reports Router observations.

\subsection{Metrics}\label{sec:8.2}

The pilot measures agreement on Dimension matching, on positions after flipping reversed scales, and on 0--3 relevance grades, where per-query kappa below 0.4 triggers re-adjudication; it also measures judgment noise and, against the pilot's own assignment, the errors, saturation and confidence of the decision calls. These pilot quantities are descriptive; Appendix B defines them and the retrieval metrics of Section 8.5, with each one's population and source.

\subsection{Results}\label{sec:8.3}

\textbf{Dimension authoring.} Table 5 compares five independently authored, post hoc matched pairs; it measures discovery correspondence and positions conditional on joint relevance. PEP has two same-axis pairs, four of five authored Dimensions (Dice share 0.8), with Spearman correlation on positions 0.94 above Section 5.5's 0.7 Measurable bar and 0.67 just below it. Enron has one same-axis pair at 0.48 and two partial, reversed pairs at 0.83 and 0.56 (Dice share 0.4). All cover at least 20\% for both annotators, though Enron coverage is 0.33--0.50 versus 0.97. Spearman correlation on positions over jointly relevant Claims is 0.48 to 0.94, while applicability alpha across pairs is $-0.37$ to $0.20$. Covered's floor is the least-settled validation setting. Appendix B defines Dice share, coverage and both position statistics.

{\small
\setlength{\tabcolsep}{3pt}
\setlength{\LTpre}{6pt}
\begin{longtable}{@{}>{\raggedright\arraybackslash}p{0.098\textwidth}>{\raggedright\arraybackslash}p{0.080\textwidth}>{\raggedright\arraybackslash}p{0.098\textwidth}>{\raggedright\arraybackslash}p{0.125\textwidth}>{\raggedright\arraybackslash}p{0.125\textwidth}>{\raggedright\arraybackslash}p{0.116\textwidth}>{\raggedright\arraybackslash}p{0.116\textwidth}>{\raggedright\arraybackslash}p{0.134\textwidth}@{}}
\multicolumn{8}{@{}p{\textwidth}@{}}{\normalsize Table 5. Dimension authoring agreement: the five matched pairs of Dimensions that annotators A and B authored independently, with positions compared over the Claims both marked relevant. Dice share is per Domain and repeats for its pairs; alpha is Krippendorff alpha on positions. Pair abbreviations are DS, decision settlement; AM, adoption maturity; MC, mechanism continuity; DAO, direct-access openness; DAV, direct-access viability; SB, switching burden; and FC, financing constraint. Definitions in Appendix B.}\\[4pt]
\toprule
Domain & Pair & Axis match & Dice & alpha & Spearman & Coverage (A / B) & n both relevant \\
\midrule
\endfirsthead
\multicolumn{8}{@{}>{\raggedright\arraybackslash}p{\textwidth}@{}}{\normalsize\bfseries Table 5. Dimension authoring agreement (continued).}\\[4pt]
\toprule
Domain & Pair & Axis match & Dice & alpha & Spearman & Coverage (A / B) & n both relevant \\
\midrule
\endhead
\bottomrule
\endlastfoot
PEP & DS / AM & same & 0.8 & 0.86 & 0.94 & 0.93 / 1.00 & 28 \\
PEP & MC / MC & same & 0.8 & 0.68 & 0.67 & 0.80 / 0.97 & 24 \\
Enron & DAO / DAV & same & 0.4 & 0.38 & 0.48 & 0.47 / 0.97 & 14 \\
Enron & SB / DAV & partial, reversed & 0.4 & 0.74 & 0.83 & 0.33 / 0.97 & 10 \\
Enron & FC / DAV & partial, reversed & 0.4 & 0.38 & 0.56 & 0.50 / 0.97 & 15 \\
\end{longtable}
}

\textbf{Relevance grading.} On PEP and Enron respectively, Cohen kappa on grades is 0.52 and 0.32, exact agreement 0.64 and 0.50, and ordinal alpha on grades 0.78 and 0.70 (Figure 9): ordering agrees more than exact grades, less on the informal corpus. Six of twelve queries fall below the 0.4 trigger (2/6 PEP and 4/6 Enron), with another PEP query exactly at it.

\begin{figure}[H]
\centering
\includegraphics[width=5.25in]{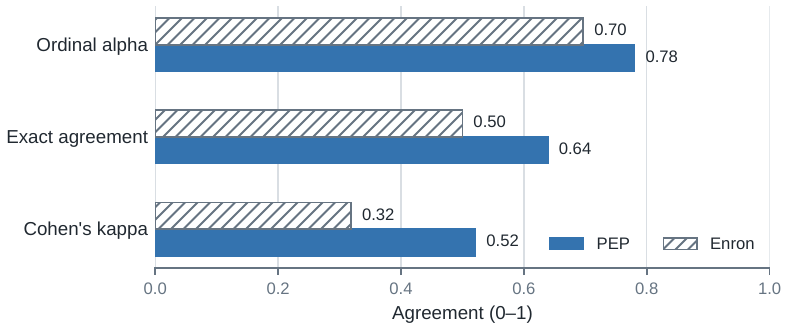}
\caption{Relevance-grade agreement between annotators on the 36 query-Claim pairs per Domain: Cohen kappa, exact agreement and ordinal alpha on grades, for PEP and Enron. Definitions in Appendix B.}
\label{fig:agreement}
\end{figure}

\textbf{Judgment noise.} Table 6 reports rough spreads from 6 queries and 30 Claims per Domain. Our registered sample sizes used a pre-pilot per-query nDCG@10-difference SD of 0.15.

{\small
\setlength{\tabcolsep}{3pt}
\setlength{\LTpre}{6pt}
\begin{longtable}{@{}>{\raggedright\arraybackslash}p{0.300\textwidth}>{\raggedright\arraybackslash}p{0.140\textwidth}>{\raggedright\arraybackslash}p{0.140\textwidth}>{\raggedright\arraybackslash}p{0.370\textwidth}@{}}
\multicolumn{4}{@{}p{\textwidth}@{}}{\normalsize Table 6. Judgment noise on the two Domains, pilot observations only. Definitions in Appendix B.}\\[4pt]
\toprule
Measure & PEP & Enron & What it describes \\
\midrule
\endfirsthead
\multicolumn{4}{@{}>{\raggedright\arraybackslash}p{\textwidth}@{}}{\normalsize\bfseries Table 6. Judgment noise on the two Domains, pilot observations only (continued).}\\[4pt]
\toprule
Measure & PEP & Enron & What it describes \\
\midrule
\endhead
\bottomrule
\endlastfoot
SD across queries of the nDCG difference between annotators & 0.063 & 0.086 & the judgment part of the per-query variance in the registered retrieval tests \\*
Pooled per-item grade SD (0-3 scale) & 0.471 & 0.540 & relevance-grade annotation \\*
Pooled per-item position SD, range over pairs & 0.126 to 0.220 & 0.154 to 0.256 & position and projection agreement \\
\end{longtable}
}

\textbf{Decision calls.} Against the non-independent pilot assignment, Table 7 shows no Signal errors and at most two query errors. Forced choice yielded 18/24 saturated answers for jev and none for clef; per-Domain yes/no yielded none among 24 query or 354 Signal answers for either, supporting separate yes/no probabilities followed by $\theta_D$.

{\small
\setlength{\tabcolsep}{3pt}
\setlength{\LTpre}{6pt}
\begin{longtable}{@{}>{\raggedright\arraybackslash}p{0.160\textwidth}>{\raggedright\arraybackslash}p{0.190\textwidth}>{\raggedright\arraybackslash}p{0.240\textwidth}>{\raggedright\arraybackslash}p{0.170\textwidth}>{\raggedright\arraybackslash}p{0.180\textwidth}@{}}
\multicolumn{5}{@{}p{\textwidth}@{}}{\normalsize Table 7. The Domain-classifier decision-call runs: two models and two call shapes over the pilot's 12 queries and 177 Signals. Argmax errors count calls whose highest-probability Domain differs from the pilot's assignment; saturated counts returned probabilities of exactly 0 or 1.}\\[4pt]
\toprule
Model & Shape & Items & Argmax errors & Saturated \\
\midrule
\endfirsthead
\multicolumn{5}{@{}>{\raggedright\arraybackslash}p{\textwidth}@{}}{\normalsize\bfseries Table 7. The Domain-classifier decision-call runs (continued).}\\[4pt]
\toprule
Model & Shape & Items & Argmax errors & Saturated \\
\midrule
\endhead
\bottomrule
\endlastfoot
clef & forced two-way choice & 12 queries & 1 & 0 of 24 \\
jev & forced two-way choice & 12 queries & 2 & 18 of 24 \\
clef & per-Domain yes/no & 12 queries & 1 & 0 of 24 \\
jev & per-Domain yes/no & 12 queries & 1 & 0 of 24 \\
clef & per-Domain yes/no & 177 Signals & 0 & 0 of 354 \\
jev & per-Domain yes/no & 177 Signals & 0 & 0 of 354 \\
\end{longtable}
}

\textbf{Calibration.} On the five-item calibration set the blind annotator reached kappa 0.667 against the key; another annotator's sheet was not blind and matches the key, so the annotator-against-annotator kappa, also 0.667, repeats that comparison.

\subsection{What the registered plan tests}\label{sec:8.4}

Before evaluation data existed, our public plan fixed baseline, metric, test construction, threshold, sample size, stopping rule and binding drop condition for nineteen components [\citelink{cite.ref35}{35}]; bars, sizes and stopping rules live in our registration.

We tune nothing on held-out splits. For retrieval we compare hybrid scoring with and without Dimensions against permutation, Signals-only, unscoped and GraphRAG-style graph arms [\citelink{cite.ref17}{17}]. For standing we compare branchwise contradiction precision/recall with exhaustive, reference-Labeller and similarity-only checks, and propagation with global relabelling, minimal-change retraction and contest latency. Frozen-log replays test compounding and staleness with staleness probes and a Graphiti-style arm [\citelink{cite.ref42}{42}]. Other tests compare discovered with human Dimensions, candidate with reference Projectors, and Router/classifier with majority and embedding baselines; Router calibration reports both misroute rates and classifier $\theta_D$ is validation-set.

Retrieval tests one interface to persisted Claims, Domain scope, standing and Dimensions; standing and replay test revision. Roles require a qualifying model; failed qualification drops the dependent component in the form the plan states, and a component's drop condition removes it.

\subsection{The retrieval test, as scored}\label{sec:8.5}

\textbf{As run.} Three workspaces contributed 91, 88 and 93 held-out general queries, 272 pooled, to registered arms (i)--(viii); facet-only (ix) is post hoc. Primary nDCG@10, secondary Recall@50 and exploratory MRR appear in Table 8. The run record was accessible to the authors for a single session: values are described as recovered when present there, computed when derived from them, and post hoc when unregistered.

{\small
\setlength{\tabcolsep}{3pt}
\setlength{\LTpre}{6pt}
\renewcommand{\arraystretch}{1.3}
\begin{longtable}{@{}>{\raggedright\arraybackslash}p{0.410\textwidth}>{\raggedright\arraybackslash}p{0.200\textwidth}>{\raggedright\arraybackslash}p{0.170\textwidth}>{\raggedright\arraybackslash}p{0.170\textwidth}@{}}
\multicolumn{4}{@{}p{\textwidth}@{}}{\normalsize Table 8. The nine arms on the 272 pooled held-out queries: mean nDCG@10 with the per-query standard deviation in parentheses, mean Recall@50 and mean MRR. We added arm (ix) post hoc. Definitions in Appendix B.}\\[4pt]
\toprule
Arm & nDCG@10 (SD) & Recall@50 & MRR \\
\midrule
\endfirsthead
\multicolumn{4}{@{}>{\raggedright\arraybackslash}p{\textwidth}@{}}{\normalsize\bfseries Table 8. The nine arms on the 272 pooled held-out queries (continued).}\\[4pt]
\toprule
Arm & nDCG@10 (SD) & Recall@50 & MRR \\
\midrule
\endhead
\bottomrule
\endlastfoot
(i) hybrid baseline & 0.421 (0.199) & 0.609 & 0.400 \\
(ii) hybrid plus standing penalty & 0.453 (0.193) & 0.627 & 0.417 \\
(iii) full reranker & 0.481 (0.199) & 0.656 & 0.441 \\
(iv) permutation control & 0.456 (0.213) & 0.637 & 0.422 \\
(v) Signals-only & 0.400 (0.222) & 0.582 & 0.378 \\
(vi) global retrieval & 0.372 (0.207) & 0.562 & 0.349 \\
(vii) graph-augmented & 0.450 (0.219) & 0.643 & 0.428 \\
(viii) graph-augmented on Signals-only & 0.428 (0.194) & 0.605 & 0.397 \\
(ix) facet-only, post hoc & 0.464 (0.224) & 0.639 & 0.426 \\
\end{longtable}
}

\begin{figure}[htbp]
\centering
\includegraphics[width=5.25in]{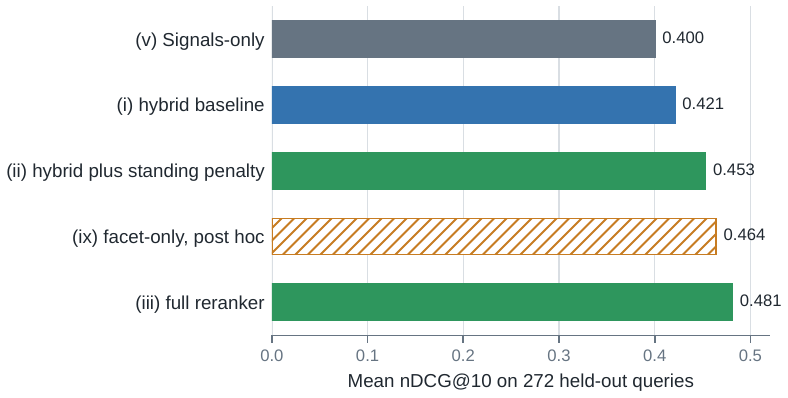}
\caption{Retrieval ablation comparison on the 272 pooled held-out queries, using the mean nDCG@10 values reported in Table 8; higher is better. The bars show absolute scores for each arm. Facet-only is a post hoc attribution arm. Statistical interpretation, registered contrasts and uncertainty remain in Table 9. Definitions in Appendix B.}
\label{fig:retrieval-ablation}
\end{figure}

\textbf{Gains against registered bars.} Table 9 preserves recovered intervals and p-values. The primary Dimension gain is 0.028 under the one-sided 0.03-per-look rule. Five secondary clauses form one Holm family: standing; one system clause requiring both contrasts against (v) and (viii) to pass, scored on 251 of 272 queries under the registered ``about 250'' sizing at strictest level 0.01 within the pooled 270; per-workspace Dimension non-inferiority; Domain scoping; and graph gain. The recovered worst non-inferiority lower bound, $-0.018$, exceeds the $-0.03$ margin, but we withhold its inconsistent per-workspace breakdown, so this clause rests on that bound alone. Because the record's decision column was empty, we computed Holm step-down [\citelink{cite.ref25}{25}]: all four p-valued clauses pass family level 0.05, and non-inferiority passes on its bound. Permutation retained 0.11 of the Dimension gain, below 0.5. Outside the family, HyDE fired on 58 queries versus 56 required and passed. On the recovered representational criteria, every efficacy bar we report clears: Dimensions and permutation pass, standing and Domain scoping remain, graph augmentation does not match the system, and escalation remains.

{\small
\setlength{\tabcolsep}{3pt}
\setlength{\LTpre}{6pt}
\begin{longtable}{@{}>{\raggedright\arraybackslash}p{0.098\textwidth}>{\raggedright\arraybackslash}p{0.080\textwidth}>{\raggedright\arraybackslash}p{0.098\textwidth}>{\raggedright\arraybackslash}p{0.125\textwidth}>{\raggedright\arraybackslash}p{0.125\textwidth}>{\raggedright\arraybackslash}p{0.116\textwidth}>{\raggedright\arraybackslash}p{0.116\textwidth}>{\raggedright\arraybackslash}p{0.134\textwidth}@{}}
\multicolumn{8}{@{}p{\textwidth}@{}}{\normalsize Table 9. The gains on the pooled held-out queries, nDCG@10: the arms compared, the query count, the difference of means, the interval and p-value as recovered from the record, the registered bar and the outcome. The Dimension gain's interval is not centred on its difference and we could not re-derive it. Definitions in Appendix B.}\\[4pt]
\toprule
Gain & Arms & n & Difference & Interval & p & Bar & Outcome \\
\midrule
\endfirsthead
\multicolumn{8}{@{}>{\raggedright\arraybackslash}p{\textwidth}@{}}{\normalsize\bfseries Table 9. The gains on the pooled held-out queries, nDCG@10 (continued).}\\[4pt]
\toprule
Gain & Arms & n & Difference & Interval & p & Bar & Outcome \\
\midrule
\endhead
\bottomrule
\endlastfoot
Dimension (primary) & (iii) minus (ii) & 272 & 0.028 & 0.014 to 0.056 & 0.002 & positive and significant, one-sided 0.03 per look & passes \\
Standing & (ii) minus (i) & 272 & 0.032 & 0.004 to 0.052 & 0.011 & Holm family & passes \\
System, Signals-only & (iii) minus (v) & 251 & 0.081 & 0.058 to 0.112 & $< 0.001$ & Holm family, both must pass & passes \\
System, graph-augmented on Signals-only & (iii) minus (viii) & 251 & 0.053 & 0.033 to 0.083 & $< 0.001$ & the same clause & passes \\
Domain scoping & (i) minus (vi) & 272 & 0.049 & 0.024 to 0.076 & $< 0.001$ & Holm family & passes \\
Graph & (iii) minus (vii) & 272 & 0.031 & 0.009 to 0.055 & 0.004 & Holm family & passes \\
Claims-in-index & (i) minus (v) & 272 & 0.021 & 0.001 to 0.043 & 0.020 & none, reported only & reported \\
HyDE escalation & (iii) with against without, on firing queries & 58 & 0.061 & 0.018 to 0.104 & 0.003 & 56 firing queries; positive and significant, one-sided 0.05 & passes \\
\end{longtable}
}

\textbf{Attribution.} Dimension gain (iii)--(ii) holds Claim persistence and the standing penalty constant; Claims-in-index (i)--(v), with no registered bar, shows only that indexing Claims improves retrieval over Signals. Neither tests dependency tracking, maintenance or upkeep. The post hoc arm asks what position adds beyond relevance.

\textbf{Post hoc facet arm.} Arm (ix) matches (iii) except that each Dimension contributes only Claim relevance: it fixes the positional factor at 1, omits query projection and refits coefficients under the same validation procedure and tuning limit. Over (ii) it gains 0.011 (interval $-0.006$ to 0.036); (iii) gains 0.017 over it (0.000 to 0.040); retention is 0.39 against the proposed 0.5 maximum. The position-specific increment is post hoc and not significant.

\textbf{Evaluation scope and limitations.} Eight registered retrieval arms were scored on 272 held-out queries, with a ninth facet-only arm evaluated post hoc. Pilot observations provide descriptive evidence, not held-out validation. Supporting records for bootstrap estimates, positive-control verification, sequential looks and original Holm decisions were unavailable. The reported retrieval gains are therefore preliminary. Truth maintenance under changing evidence, sustained organizational reuse and organizational instinct were not evaluated. The preregistered evaluation plan, available execution records, and pilot materials are publicly accessible in the accompanying repository [\citelink{cite.ref35}{35}].

\section{Conclusion}\label{sec:9}

The proposed shared organizational cortex composes attributable Signals, contextual Domains, Claims with justifications and assumptions, ordered Dimension positions, and standing maintained through justification, conflict and supersession dependencies.

Our registered test reports a 0.028 nDCG@10 Dimension gain ($p=0.002$) and positive significant gains over Signals-only and graph-augmented retrieval. Distinct Claims-in-index, standing, Domain-scoping, Dimension and facet-only contrasts attribute retrieval effects, but our post hoc analysis establishes no significant position-specific increment. Cohen kappa on grades is 0.52 and 0.32; Dimension-applicability alpha ranges from $-0.37$ to $0.20$. This is our preliminary representational evidence.

Open problems include maintenance under change, contradiction recall, Extractor recall, Dimension applicability, governance over who may read or accept a Claim, assumption-based contexts for unevidenced preconditions, conflict chains under argumentation semantics and integral Dimension groups. The contribution is a formal foundation for maintained interpretation and cross-context reuse, with one evaluated consequence rather than a completed account of organizational intelligence.

\clearpage
\appendix
\section{Retrieval Execution Details}\label{sec:A}

\subsection{Router}\label{sec:A.1}

The optional Router is a decision-model call over retrieve, reason and author, given request, workspace and interaction context. It selects the highest-probability label, escalating to the next more constructive mode when that probability is below a validation-set threshold. Domain classification remains semantically separate even if one call combines them.

Over 12 queries, clef and jev had mean selected-option confidence 0.49 and 0.72 and saturation 0/36 and 2/36. These test call shape.

\subsection{HyDE escalation}\label{sec:A.2}

Hypothetical Document Embeddings (HyDE; [\citelink{cite.ref19}{19}]) optionally embed at most one hypothetical answer when retrieval confidence is low. HyDE embeds one hypothetical answer as a retrieval key; it is a retrieval implementation.

In our registered test, HyDE fired on 58 held-out queries versus 56 required and improved nDCG@10 by 0.061 (interval 0.018--0.104, $p=0.003$); we retain the bounded escalation under its registered bar, which bears nothing on maintenance or organizational instinct.

\section{Metric definitions}\label{sec:B}

This appendix defines the metrics and statistics that Section 8 and Appendix A report. The annotation quantities compare annotators A and B on two Domains, PEP and Enron; we computed them with the pilot scorer released with the pilot materials [\citelink{cite.ref35}{35}], and the decision-call quantities from the logged calls. Each Domain has 30 Claims, over which positions and coverage are taken, and 6 queries with 6 candidate Claims each, its 36 graded query--Claim pairs. Grades run from 0 to 3: 3, the Claim answers the query or is required to answer it; 2, it answers part of it; 1, it is on the query's topic but does not help answer it; 0, otherwise. An annotator positions each Claim on each Dimension they authored, or marks it NR when the Claim does not concern that Dimension; a scalar position lies in $[0, 1]$, and level $i$ of $L$ ordinal levels becomes $(i-1)/(L-1)$. After submission, the Dimension sets were matched from definitions and prototypes into same-axis, partial or unmatched pairs; in a reversed pair one annotator's high end is the other's low end, and B's positions $x$ become $1-x$ before comparison. Retrieval quantities are taken over the 272 pooled held-out queries, 91, 88 and 93 from the three workspaces, for arms (i) to (ix) as Table 8 names them.

{\small
\setlength{\tabcolsep}{3pt}
\setlength{\LTpre}{6pt}
\begin{longtable}{@{}>{\raggedright\arraybackslash}p{0.170\textwidth}>{\raggedright\arraybackslash}p{0.250\textwidth}>{\raggedright\arraybackslash}p{0.290\textwidth}>{\raggedright\arraybackslash}p{0.240\textwidth}@{}}
\multicolumn{4}{@{}p{\textwidth}@{}}{\normalsize Table 10. Metric definitions: one name per quantity, its definition, the population it is computed over, and where the paper reports it. Formulas marked ``below'' follow the table.}\\[4pt]
\toprule
Quantity & Definition & Computed over & Reported in \\
\midrule
\endfirsthead
\multicolumn{4}{@{}>{\raggedright\arraybackslash}p{\textwidth}@{}}{\normalsize\bfseries Table 10. Metric definitions (continued).}\\[4pt]
\toprule
Quantity & Definition & Computed over & Reported in \\
\midrule
\endhead
\bottomrule
\endlastfoot
Dice share of matched Dimensions & $(m_A+m_B)/(n_A+n_B)$; below & The Dimensions authored on both sides in each Domain; only same-axis pairs count & Table 5; Section 8.3 \\
Krippendorff alpha on positions & $1-D_o/D_e$; level below & Each matched pair, over the jointly relevant Claims & Table 5 \\
Spearman correlation on positions & Pearson correlation of the two rank sets, ties at mean rank & Each matched pair, over the jointly relevant Claims & Table 5; Section 8.3 \\
Coverage & Share of the Claims positioned, not marked NR & Each annotator and Dimension, over the Domain's 30 Claims & Table 5; Sections 5.4 and 8.3 \\
Cohen kappa on grades & Unweighted; below & The 36 graded pairs per Domain & Section 8.3; Figure 9; Section 9 \\
Per-query kappa & Cohen kappa within one query; below 0.4 triggers re-adjudication & Each query's 6 candidates; 12 queries & Section 8.3 \\
Exact agreement & Share of pairs given the same grade & The 36 graded pairs per Domain & Section 8.3; Figure 9 \\
Ordinal alpha on grades & Krippendorff alpha, ordinal level, grades 0 to 3 & The 36 graded pairs per Domain & Section 8.3; Figure 9 \\
Calibration kappa & Cohen kappa on grades, an annotator against the key or against another annotator & Five calibration items & Section 8.3 \\
Across-query SD of the annotators' nDCG difference & Sample SD, divisor $n-1$, of $\mathrm{nDCG}_A-\mathrm{nDCG}_B$ & The 6 queries per Domain; pilot nDCG over the candidates in the kit's fixed order & Table 6 \\
Pooled per-item grade SD & Root mean square of $\lvert a-b \rvert/\sqrt{2}$; below & The 36 graded pairs per Domain & Table 6 \\
Pooled per-item position SD & As for grades, on normalized positions & Each matched pair, over the jointly relevant Claims; range over pairs & Table 6 \\
Argmax errors & Calls whose top Domain differs from the pilot's assignment & 12 queries or 177 Signals, per model and call shape & Table 7; Section 8.3 \\
Saturation & Returned probabilities of exactly 0 or 1 & All returned: two per item for Domain calls (24 or 354), three per Router query (36) & Table 7; Appendix A.1 \\
Selected-option confidence & Mean confidence a choice-shaped call returns; below & 12 Router queries per model & Appendix A.1 \\
nDCG@10 & Mean of DCG@10 over ideal DCG@10; below & The pooled held-out set, per arm & Tables 8 and 9; Figure 10; Sections 8.5 and 9 \\
Recall@50 & Mean share of judged relevant items in the top 50 & The pooled held-out set, per arm & Table 8 \\
MRR & Mean reciprocal rank of the first relevant item; exploratory & The pooled held-out set, per arm & Table 8 \\
Gain & First arm's mean nDCG@10 minus the second's & The n of each Table 9 row & Table 9; Sections 8.5 and 9 \\
Interval and p-value & One-sided paired bootstrap over queries; below & The queries of each gain & Table 9; Section 8.5; Appendix A.2 \\
Holm step-down & Sorted p-values tested in turn, rank $j$ at $0.05/(m-j+1)$ with $m=5$, until one fails & The five secondary clauses, four with p-values & Section 8.5; Table 9 \\
Per-workspace non-inferiority lower bound & Lower confidence bound on one workspace's Dimension gain; margin $-0.03$ & 91, 88 and 93 queries & Section 8.5 \\
Retention & An arm's gain over (ii) divided by the Dimension gain; below & Arms (iv) and (ix), 272 queries & Section 8.5 \\
HyDE firing count & Held-out queries on which HyDE escalation fired; 56 required & The 272 queries & Section 8.5; Table 9; Appendix A.2 \\
\end{longtable}
}

nDCG follows Järvelin and Kekäläinen [\citelink{cite.ref27}{27}]. With $\ell_i$ the grade of the item at rank $i$ and a gain mapping $g$,

\[
\mathrm{DCG@}k = \sum_{i=1}^{k} \frac{g(\ell_i)}{\log_2(i+1)}, \qquad \mathrm{nDCG@}k = \frac{\mathrm{DCG@}k}{\mathrm{IDCG@}k},
\]

where $\mathrm{IDCG@}k$ is the DCG of the same items in ideal order. The pilot uses $g(\ell) = 2^{\ell} - 1$ over all six candidates of a query. For the retrieval runs we state the standard definition with $k = 10$; their gain mapping and treatment of unjudged items are not recorded. Recall@50 counts relevant items among the judged ones, and MRR averages the reciprocal rank of the first relevant item [\citelink{cite.ref49}{49}]; for both, the grade at which an item counts as relevant is not recorded.

Krippendorff alpha is $\alpha = 1 - D_o/D_e$, the observed disagreement between the values given to the same unit over the disagreement expected when all pooled values are paired by chance [\citelink{cite.ref32}{32}]; units with one value drop out. Positions use the interval level, with squared differences, except one Enron pair whose Dimensions are both ordinal with the same number of levels; it uses the ordinal level with every level of the scale in the value domain. Grades use the ordinal level over 0 to 3. Cohen kappa [\citelink{cite.ref10}{10}] treats each grade as its own category:

\[
\kappa = \frac{p_o - p_e}{1 - p_e}, \qquad p_e = \sum_{g} p_A(g)\, p_B(g),
\]

where $p_o$ is the share of items given the same grade and $p_A(g)$ and $p_B(g)$ the shares each annotator gave grade $g$. Judgments $a_i$ and $b_i$ of item $i$ have sample SD $\lvert a_i - b_i \rvert/\sqrt{2}$, pooled over $N$ items as $s = \sqrt{\sum_{i=1}^{N} (a_i - b_i)^2/(2N)}$. The Dice share and the retention of an arm $x$, with $\bar{s}$ an arm's mean nDCG@10, are

\[
\mathrm{Dice} = \frac{m_A + m_B}{n_A + n_B}, \qquad R(x) = \frac{\bar{s}_x - \bar{s}_{\mathrm{(ii)}}}{\bar{s}_{\mathrm{(iii)}} - \bar{s}_{\mathrm{(ii)}}},
\]

where $n_A$ and $n_B$ count the Dimensions each annotator authored in the Domain and $m_A$ and $m_B$ those in a same-axis pair. With one-to-one pairs the Dice share is the coefficient of Dice [\citelink{cite.ref14}{14}], $2m/(n_A + n_B)$ for $m$ pairs: $(2+2)/5 = 0.8$ for PEP and $(1+1)/5 = 0.4$ for Enron. Table 8 reproduces both reported retentions, $0.003/0.028 = 0.107$ for the permutation control (iv) and $0.011/0.028 = 0.393$ for the facet-only arm (ix); the plan requires the first to be at most 0.5.

Intervals and p-values come from a one-sided paired bootstrap over queries [\citelink{cite.ref45}{45}]: queries are resampled with replacement, each keeping its scores under both arms, and the gain is recomputed on each resample. The confidence level and the number of resamples are not recorded. The primary Dimension gain follows the plan's two-look rule: a positive but non-significant first result extends the sample once by the same size and the pooled sample is tested again, each look at a nominal one-sided 0.03, holding the overall type I error near 0.05 [\citelink{cite.ref41}{41}]; which look decided the reported test is not recorded, since the records of the looks are unavailable (Section 8.5).

Holm step-down runs over five secondary clauses: standing, the system clause, per-workspace non-inferiority, Domain scoping and the graph gain. The system clause passes only if both its contrasts pass at its level; non-inferiority is judged by its lower bound, which must exceed $-0.03$ in every workspace, and the level at which the reported $-0.018$ was taken is not recorded. The registered plan is not consistent on the graph gain: one list in it and its amendments record that gain as reported only, while its statement of the retrieval test places it in the Holm family; we follow that statement. Selected-option confidence is the confidence value a decision call returns with its choice; it is not the probability of the chosen option, and how the call computes it is not recorded.

\section{Declarations}\label{sec:C}

\textbf{Funding.} Robin Inc. funded this work.

\textbf{Author contributions.} Andrew Miracle conceptualized the architecture and study and conducted the research and analyses. Phyl Georgiou and Christine Okoth jointly supervised the project. Ese Monday reviewed the manuscript, contributed pilot annotation, and prepared pilot reports and visualizations, including figures and diagrams.

\textbf{Ethics and AI assistance.} The study analysed public documents and an existing public email corpus. Author-researchers supplied pilot annotation judgments; no external human participants were recruited. The pilot also collected outputs from language models for the Domain-classifier and Router call-shape observations; these were treated as system measurements, not human-participant data. Generative-AI tools assisted with structured entry of author judgments, editorial revision, and preparation of manuscript and build artifacts.

\textbf{Competing interests.} Andrew Miracle, Phyl Georgiou and Christine Okoth are employed by or hold interests in Robin Inc., which funded this work and develops systems based on the architecture described.

\end{document}